\documentclass{article}

\usepackage{microtype}
\usepackage{graphicx}
\usepackage{subfigure}
\usepackage{booktabs} 
\usepackage{multirow}
\usepackage{lmodern}
\usepackage{bm}
\usepackage{amsmath}
\usepackage{bbm}
\usepackage{dsfont}
\usepackage{hyperref}

\usepackage[accepted]{icml2024}

\usepackage{amsmath}
\usepackage{amssymb}
\usepackage{mathtools}
\usepackage{amsthm}

\usepackage[capitalize,noabbrev]{cleveref}

\theoremstyle{plain}

\theoremstyle{definition}

\theoremstyle{remark}

\usepackage[textsize=tiny]{todonotes}

\icmltitlerunning{OMH: Structured Sparsity via Optimally Matched Hierarchy for Unsupervised Semantic Segmentation}

\begin{document}

\twocolumn[
\icmltitle{OMH: Structured Sparsity via Optimally Matched Hierarchy \\ for Unsupervised Semantic Segmentation}




\begin{icmlauthorlist}
\icmlauthor{Baran Ozaydin}{yyy}
\icmlauthor{Tong Zhang}{yyy}
\icmlauthor{Deblina Bhattacharjee}{yyy}
\icmlauthor{Sabine Süsstrunk}{yyy}
\icmlauthor{Mathieu Salzmann}{yyy}
\end{icmlauthorlist}

\icmlaffiliation{yyy}{Computer and Communication Sciences, EPFL, Lausanne, Switzerland}

\icmlcorrespondingauthor{firstname lastname}{firstname.lastname@epfl.ch}


\vskip 0.3in
]



\printAffiliationsAndNotice{}  


\begin{abstract}
Unsupervised Semantic Segmentation (USS) involves segmenting images without relying on predefined labels, aiming to alleviate the burden of extensive human labeling. Existing methods utilize features generated by self-supervised models and specific priors for clustering. However, their clustering objectives are not involved in the optimization of the features during training. Additionally, due to the lack of clear class definitions in USS, the resulting segments may not align well with the clustering objective. 
In this paper, we introduce a novel approach called Optimally Matched Hierarchy (OMH) to simultaneously address the above issues. The core of our method lies in imposing structured sparsity on the feature space, which allows the features to encode information with different levels of granularity.
The structure of this sparsity stems from our hierarchy (OMH).
To achieve this, we learn a soft but sparse hierarchy among parallel clusters through Optimal Transport.
Our OMH yields better unsupervised segmentation performance compared to existing USS methods. Our extensive experiments 
demonstrate the benefits of OMH when utilizing our differentiable paradigm. We will make our code publicly available.
\end{abstract}    
\section{Introduction}
\label{sec:intro}

\begin{figure}[ht]
\includegraphics[width=1\linewidth]{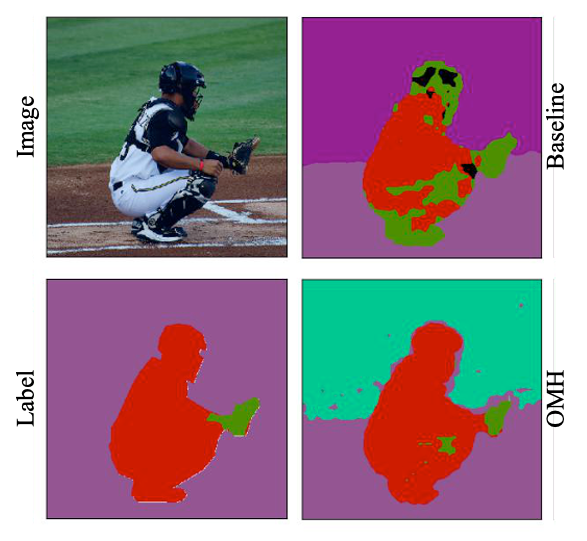}
\vspace{-0.8cm}
\caption{{\bf Our method represents classes and their parts more accurately.} Our OMH models intercluster relationships explicitly in a hierarchy and leads to better coverage of classes. Our method can link the head of the human to its body.}
\label{fig:teaser}
\vspace{-0.5cm}
\end{figure}

Semantic Segmentation (SS) aims to classify each pixel in an image into a single semantic class. The pixel-wise nature of SS makes it a dense task over the pixel space whereas assignment to a single class makes this task sparse over the classes. In the supervised scenario~\cite{long2015fcn, cheng2021maskformer, yu2022kmaxdeeplab}, these pixel-wise dense and class-wise sparse properties are reflected by the one-hot pixel-wise annotations. However, these annotations are costly to obtain, which prevents the scalability and generalizability of semantic segmentation.

This challenge has motivated the development of Unsupervised Semantic Segmentation (USS) methods, which do not require ground-truth annotations. In particular, recent works~\cite{hamilton2022stego, melas2022deepspectral, li2023acseg} leverage the progress in Self-Supervised Learning (SSL)~\cite{bardes2021vicreg, caron2021dino} to extract spatially dense features, which can then be clustered to obtain a segmentation.
While this addresses the pixel-wise density property of the SS paradigm, none of the existing USS methods have tackled the class-wise sparsity one. We attribute this to the lack of clear class definitions in the unsupervised scenario, making it difficult to identify semantic concepts. As a result, the clusters obtained by existing USS methods do not overlap with the semantic classes accurately.

\begin{figure*}[ht]
\includegraphics[width=1\linewidth]{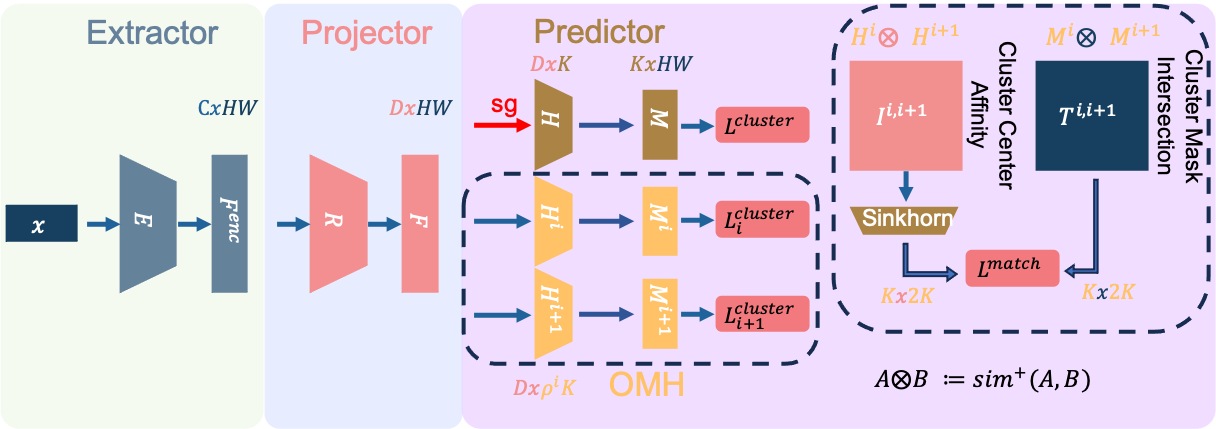}
\vspace{-0.6cm}
\caption{{\bf Overview of our method.} We impose sparsity on the features via clustering loss $L^\text{cluster}_i$ and bring a structure to this sparsity via a Wasserstein loss $L^\text{match}$. Our $L^\text{match}$ limits the intersection between clusters from different hierarchy levels using Optimal Transport~\cite{cuturi2013sinkhornot}.} 
\label{fig:model}
\vspace{-0.3cm}
\end{figure*}

In this work, we overcome these challenges by imposing a \emph{structured sparsity} in the feature space to
encode the semantically sparse nature of SS via a multi-level hierarchy. Sparsity arises from the backpropagation of the clustering loss to the previous layers (projector head) whereas structure stems from having multiple clusterings (predictor heads) that are related to each other via an Optimally Matched Hierarchy (OMH). We refer to this hierarchy as \textit{optimally matched} because it is not imposed by division or agglomeration of clusters but learned based on a soft but sparse hierarchy matrix, obtained by solving an Optimal Transport problem. The learned hierarchy matrix denotes the relationship between cluster centers (i.e. unsupervised class definitions) from different levels of the hierarchy and is then used to match the cluster activations of these levels. As a result, a hierarchical structure allows features to encode details at multiple granularities whereas the sparsity backpropagated from this hierarchy reflects the semantically sparse nature of SS. Our formalism is general and can be incorporated in many existing USS frameworks, such as STEGO~\cite{hamilton2022stego}, HP~\cite{seong2023hiddenpos}, SmooSeg~\cite{lan2023smooseg}, providing a consistent boost in accuracy.

Our contributions can be summarized as follows:
\begin{itemize}
    \item We introduce a novel approach that exploits structured semantic sparsity to achieve USS. Our proposed approach inherently encodes the semantically sparse nature of SS.
    \item To achieve this structured sparsity, we present a novel Optimally Matched Hierarchy (OMH), which is learned during the training process.
    \item This is done by formulating the hierarchical clustering as an Optimal Transport problem reflecting the semantically sparse nature of SS in an optimal transport formalism.
    \item Our OMH is general and can be exploited to boost the performance of multiple USS methods.
\end{itemize}
Our experiments demonstrate that the features learned by our method lead to state-of-the-art performance in USS on COCOStuff, Cityscapes, and Potsdam in terms of mIoU and Accuracy. Our code will be made publicly available.

\section{Related Work}
\label{sec:related}

\subsection{Self-supervised Learning}
Our method borrows concepts from previous work in Self-supervised Learning (SSL)~\cite{chen2020simclr, he2020mocov1, caron2020swav, zbontar2021barlow, bardes2021vicreg, zhu2023vcr, caron2021dino, Zhang:etal:2022, qiu2023coupling}. The aim of SSL is to learn representations that are useful for downstream tasks without ground-truth annotations. A standard approach to doing so is to compare the feature vector of different samples~\cite{chen2020simclr, he2020mocov1, grill2020byol, caron2020swav, chen2021simsiam}. 
In this context, SimCLR~\cite{chen2020simclr} uses a contrastive loss with augmentations and a large batch-size; MoCo~\cite{he2020mocov1} uses momentum contrast to stabilize the contrastive loss with a small batch size; SwAV~\cite{caron2020swav} swaps the clusters of augmented representations and uses Sinkhorn's  algorithm~\cite{sinkhorn1967sinkhorn} to obtain balanced clusters; SimSiam~\cite{chen2021simsiam} shows that a predictor without negative samples, clustering, or a moving average encoder is comparable to \cite{chen2020simclr, grill2020byol, caron2020swav}. Similarly to SwAV~\cite{caron2020swav}, our work utilizes Sinkhorn's algorithm~\cite{sinkhorn1967sinkhorn}.  In contrast to SwAV, however, we use it to compute the optimal transportation plan
between two levels of the cluster hierarchy.

A different line of research seeks to compare the sample vectors of different features~\cite{zbontar2021barlow, bardes2021vicreg, zhu2023vcr} 
In this context, Barlow Twins~\cite{zbontar2021barlow} shows that reducing the correlations of features while preserving their variance leads to comparable performance with other SSL methods. VicReg~\cite{bardes2021vicreg} and VCR~\cite{zhu2023vcr} then express the previous SSL losses in terms of variance, covariance, and invariance. 

The above-mentioned works target image-level features, which makes them ill-suited to USS, where diverse semantics and complex scenes might lead to collapse~\cite{Zhang:etal:2022, qiu2023coupling}. Dense SSL works have thus been proposed, providing better initialization for dense tasks~\cite{wang2021densecl,xie2021propagate, qiu2023coupling}. Similarly, our work aims to learn dense representations. In contrast to these approaches, we do not aim to train SSL methods from scratch for semantic segmentation but build up on generalist SSL methods and learn specialized, useful features for USS. Hence, our work is computationally less expensive than the dense SSL works.

Lastly, following the success of vision transformers (ViT)~\cite{dosovitskiy2020vit}, DINO~\cite{caron2021dino} shows that self-supervised transformers learn dense representations that contain cues of the underlying scene semantics. 

\subsection{Semantic Segmentation}
Supervised segmentation models~\cite{long2015fcn, cheng2021maskformer, kirillov2023sam} require dense and pixel-wise supervision which is expensive to obtain. 
As such, the USS methods are gaining popularity, particularly with the progress in SSL. Building on the success of Vision Transformers (ViT)~\cite{dosovitskiy2020vit} in supervised SS~\cite{cheng2021maskformer,kirillov2023sam}, many USS techniques are based on ViTs~\cite{amir2021deepvit, van2022maskdistill, melas2022deepspectral, zadaianchuk2023comus, li2023acseg, hamilton2022stego, seong2023hiddenpos}, that exploit SSL techniques and/or backbones to generate semantic predictions without ground-truth labels. In particular, DeepVIT~\cite{amir2021deepvit} shows that the dense DINO~\cite{caron2021dino} features improve the unsupervised segmentation performance. Based on this, \cite{van2022maskdistill, melas2022deepspectral, wang2022tokencut, zadaianchuk2023comus} aim to solve object-centric unsupervised semantic segmentation. MaskDistill~\cite{van2022maskdistill} extracts pseudo masks from DINO~\cite{caron2021dino}, using its CLS token and attentions, and then trains a segmentation model with these pseudo-labels. DSM~\cite{melas2022deepspectral} first applies image-wise spectral clustering~\cite{shi2000ncut, von2007spectral} to obtain foreground segments and then incorporates dataset-wise KMeans clustering~\cite{lloyd1982kmeans} to collect smaller image segments into semantically meaningful clusters. COMUS~\cite{zadaianchuk2023comus} uses pretrained and distilled saliency methods with iterative self-training. ACSeg~\cite{li2023acseg} learns a transformer decoder to perform image-wise clustering using the modularity loss. These works focus on simple scenes with few objects where the background is omitted.

For more complex scenes, STEGO~\cite{hamilton2022stego} constitutes a seminal work, proposing correspondence distillation with k-nearest neighbors~\cite{aberman2018nbb} and random images; where an extra clustering step, that is detached from training, produces the final predictions. Following this, HP~\cite{seong2023hiddenpos} incorporates local and global positives during training with contrastive learning. More recently, SmooSeg~\cite{lan2023smooseg} proposes to incorporate the distillation loss on the predictions, instead of the projected features and uses a momentum encoder for the predictor. Our OMH can be added to STEGO~\cite{hamilton2022stego}, HP~\cite{seong2023hiddenpos}, and SmooSeg~\cite{lan2023smooseg} to introduce a \textit{structured} as well as a \textit{hierarchical} sparsity that is backpropagated to their features. 
We will demonstrate the benefits of our contributions to these three methods.

\subsection{Hierarchical Clustering}
Hierarchical clustering aims to partition the data into clusters that have a structured relationship with each other, in the form of a hierarchy. Existing algorithms are either divisive or agglomerative and lead to strict partitions and hierarchical relationships~\cite{murtagh2017hier1, ran2023hier2}. Here, we utilize concepts from hierarchical clustering to introduce a \textit{soft but sparse} relationship within the hierarchy. Nevertheless, our aim is \textit{not} to propose a hierarchical clustering algorithm that gives provable results with strict hierarchies. Rather, we introduce a \textit{structural relationship} 
to obtain deep features that are, ultimately, useful for USS.
\section{Method}
\subsection{Background and Motivation}
\begin{figure*}[ht]
\begin{tabular}{cc}
    \includegraphics[width=0.97\columnwidth,trim={0 0.8cm 0 0},clip]{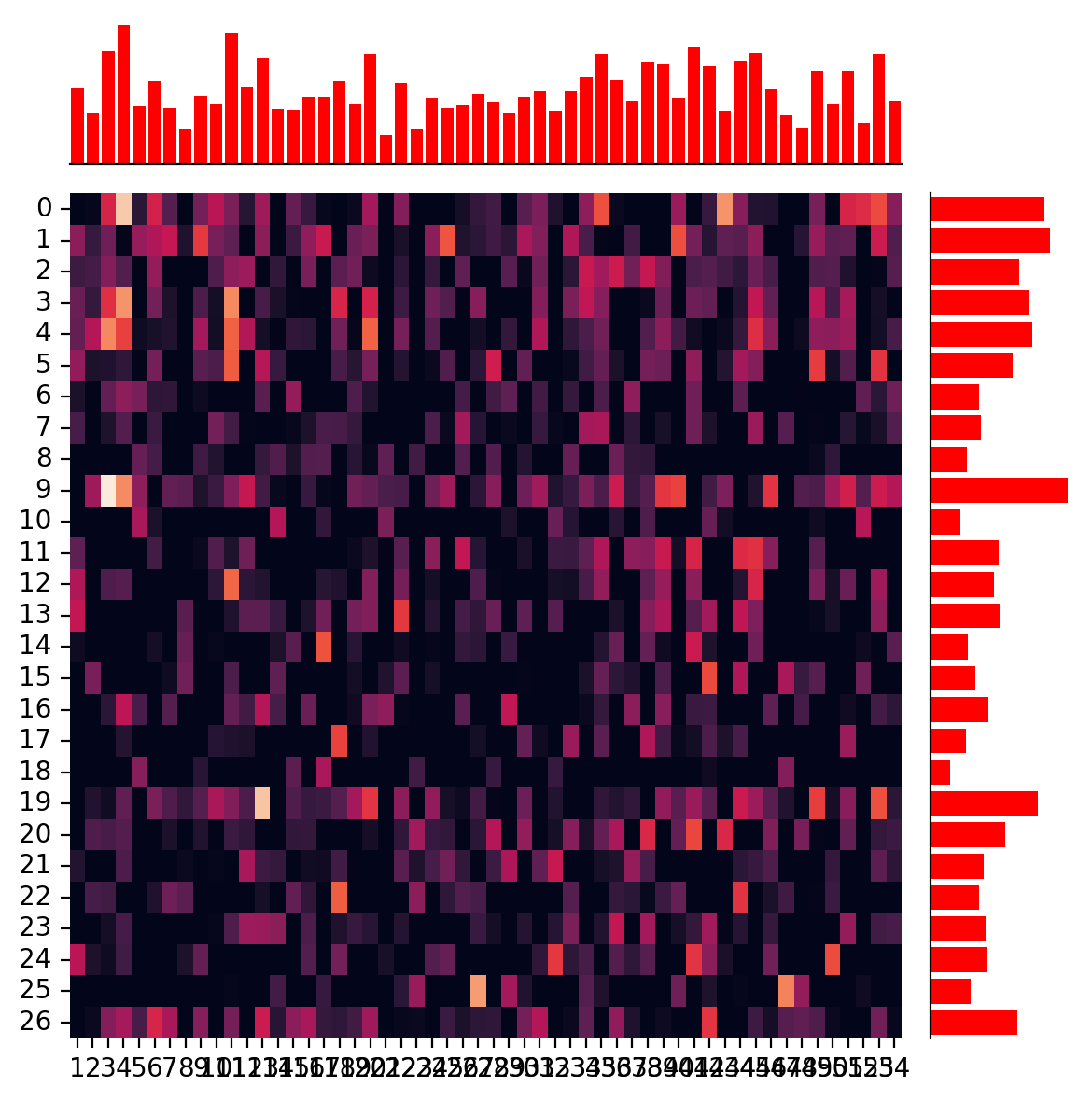} & \includegraphics[width=0.97\columnwidth,,trim={0 0.8cm 0 0},clip]{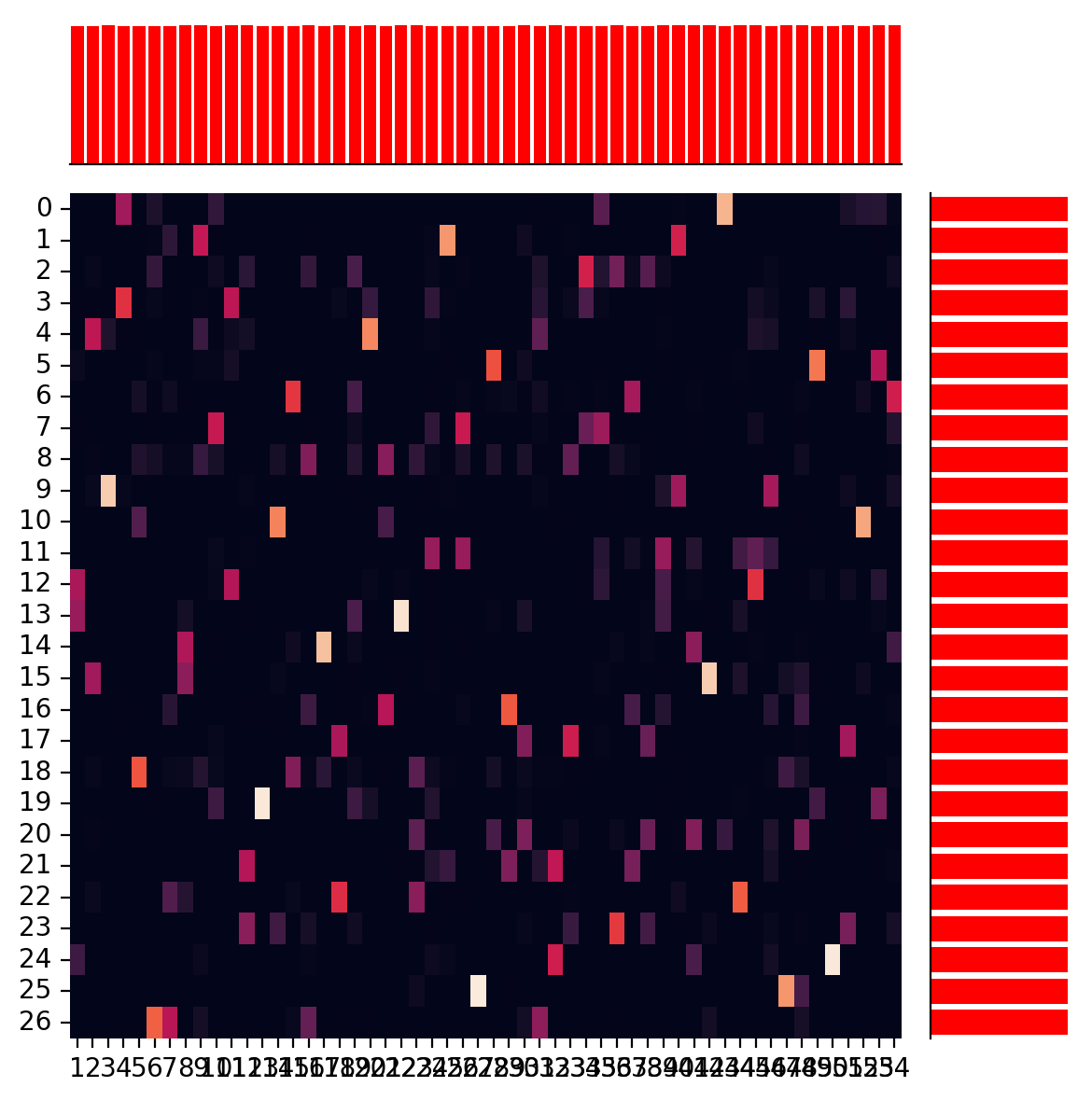} \\
     & 
\end{tabular}
\caption{{\bf Hierarchy matrix with Optimal Transport.} Optimal Transport~\cite{cuturi2013sinkhornot} leads to a sparse and balanced hierarchy. On the left, we visualize the cluster center affinities $sim^+(\bm{H}^{(i)}, \bm{H}^{(i+1)})$ which is neither sparse (many positive similarities) nor balanced (unequal row or column sums). On the right, the optimal transportation plan $\mathbf{A}^{(i,i+1)}$ is visualized. In $\mathbf{A}^{(i,i+1)}$, two lower-level clusters are mapped to one higher-level cluster, on average, but we mostly see that some higher-level clusters are mapped to three or four lower-level clusters.}
\label{fig:hier_OT}
\end{figure*}
Our approach learns a hierarchy over image regions and can be incorporated into existing USS methods, as we will show for
STEGO~\cite{hamilton2022stego}, HP~\cite{seong2023hiddenpos}, and SmooSeg~\cite{lan2023smooseg}.
In this section, we therefore provide a generic review of USS methods and their limitations, which motivated our contributions.

Given an image dataset $X \subset \mathds{R}^{3 \times H'W'}$, USS methods leverage a feature extractor backbone $E$ to compute image features. Formally, the encoder takes as input images $x, y \in X$ and outputs features $\bm{F}^{\text{enc}}_x, \bm{F}^{\text{enc}}_y$ with $C$ channels via the mapping  $E: \mathds{R}^{3 \times H'W'} \rightarrow \mathds{R}^{C \times HW}$. In many existing models, the feature extractor $E$ is a pretrained and frozen SSL backbone, such as DINO~\cite{caron2021dino}.

A projection head $R$ then processes the encoded features $\bm{F}^{\text{enc}}_x$ to produce low-dimensional features $\bm{F}_x$ via the projection $R: \mathds{R}^{C \times HW} \rightarrow \mathds{R}^{D \times HW}$. $R$ typically is a 2-layer CNN with ReLU nonlinearity~\cite{hamilton2022stego,lan2023smooseg}.

To retain the pairwise relations of the encoded features after projection, a distillation loss term (smooth loss in \cite{lan2023smooseg}) of the form
\begin{multline}
    L^{\text{distill}} = \\
    - \sum_{p,q}{(sim(\bm{F}^{\text{enc}}_x, \bm{F}^{\text{enc}}_y)_{p,q} - b) sim^+(\bm{F}_x, \bm{F}_y)_{p,q}}
\label{eq:stego_dist}
\end{multline}
is minimized, 
where $b$ is a scalar hyperparameter aiming 
to push features with smaller correlation than $b$ away from each other, $sim: \mathds{R}^{D \times HW}, \mathds{R}^{D \times MN} \rightarrow \mathds{R}^{HW \times MN}$ is the cosine similarity, $\odot$ is the element-wise product, and $p, q$ index the locations within the images $x, y$. The images $x,y$ can be identical, a kNN pair, or a random pair from the dataset $X$~\cite{hamilton2022stego}.

The extracted low-dimensional features $\bm{F}_x$ are then processed by a predictor (clustering) head $H$ that produces soft cluster affinities $\bm{M} \in \mathds{R}^{K \times HW}$ as
\begin{align}
    \bm{M}_x = sim(\bm{H}, sg(\bm{F}_x)),
    \label{eq:stego_affinity}
\end{align}
where $\bm{H} \in \mathds{R}^{K \times D}$ encodes $K$ cluster centers, and $sg$ denotes the stop gradient operation. The cluster centers (the predictor weights) are trained by minimizing
\begin{align}
    L_{\text{cluster}} = - \sum_{x, k, l}{(\bm{M}_x \hat{\bm{M}}^{\text T}_x)_{k,l}},
\label{eq:stego_cluster}
\end{align}
where $(\hat{\bm{M}})_{k,p} = \{(\bm{M})_{k,p} >= (\bm{M})_{j,p}\; \forall j\}$ is a boolean one-hot, hard cluster assignment matrix, and $k, l$ index the clusters in $\bm{M}$ and $\hat{\bm{M}}$. 

Among the functions and outputs described above, only the one-hot hard assignment matrix $\hat{\bm{M}}$ explicitly encodes both the spatial density and semantic sparsity properties of the SS task. Yet, because of the stop gradient operation, $\hat{\bm{M}}$ is not involved in the training of the features $\hat{\bm{F}_x}$. In other words, nothing encourages the learned \emph{features} $\bm{F}_x$ to explicitly encode the semantic sparsity of the SS task.

\subsection{Introducing Semantic Sparsity}
\label{parallel}
Let us now explain how we impose semantic sparsity on the low-dimensional features.
To this end, we incorporate $N$ separate predictor heads $H^{(i)}: \mathbbm{R}^{D \times HW} \rightarrow [0,1]^{(\rho^i K) \times HW}$ that are trained in parallel, for $i \in \{0,1,...,N-1\}$. Each $H^{(i)}$ yields a soft cluster assignment matrix as
\begin{align}
    \bm{M}^{(i)} = sim^+(\bm{H}^{(i)}, \bm{F}),
    \label{eq:sdh_cluster}
\end{align}
where $sim^+$ denotes the cosine similarity with negative values clipped to zero. The number of clusters $K_i$ in the predictor head $H^{(i)}$ increases with $i$ according to a multiplicative expansion factor $\rho$ as $K_i = K \times \rho^i$, e.g. 27, 54, 108 for $K=27, \rho=2, N=3$. Since each level of the parallel clusterings has a different number of clusters, they focus on different levels of detail in the features $\bm{F}$. We use these levels as the architecture of our hierarchy in SDH. Note that, compared to Eq. \ref{eq:stego_affinity}, we remove the stop gradient to impose sparsity on the features $\bm{F}$. This sparsity is imposed via multiple levels of clusterings, but it does not have a structure yet, as the clusterings are parallel and independent so far. In the next section, we constrain these clusterings to match with each other, which brings a structure to the sparsity backpropagated via Eqs.~\ref{eq:stego_cluster} and \ref{eq:sdh_cluster}.

\begin{figure*}[ht]
\begin{tabular}{cc}
    \includegraphics[width=0.97\columnwidth]{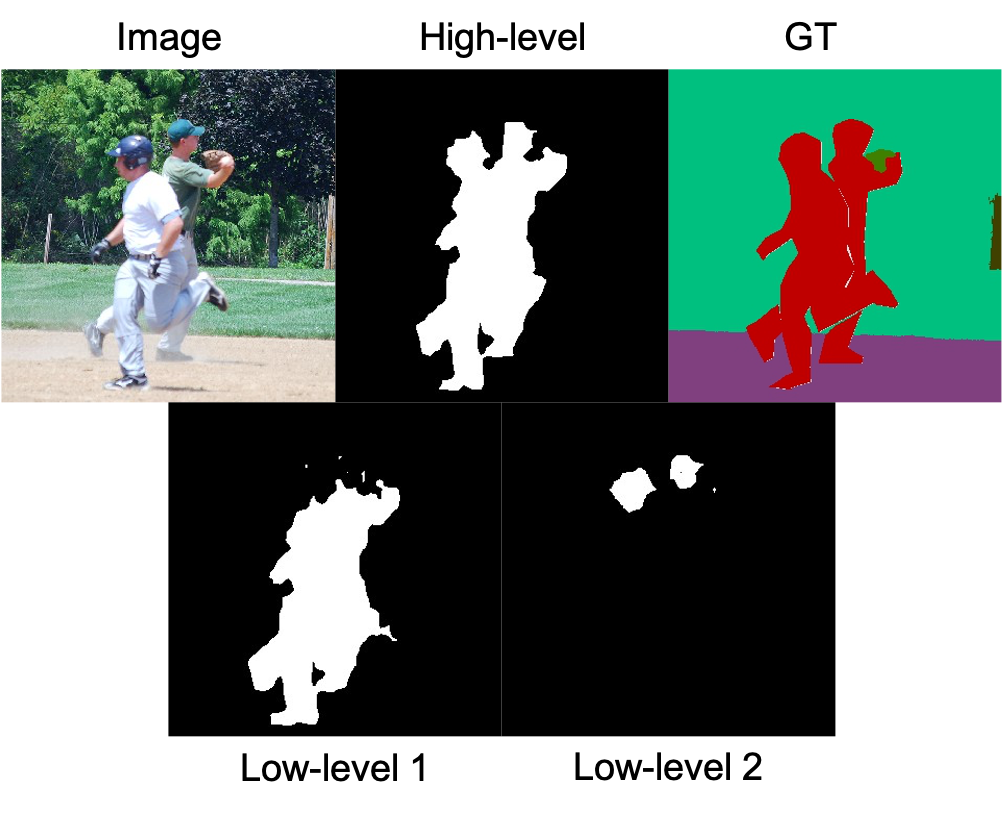} & \includegraphics[width=0.97\columnwidth]{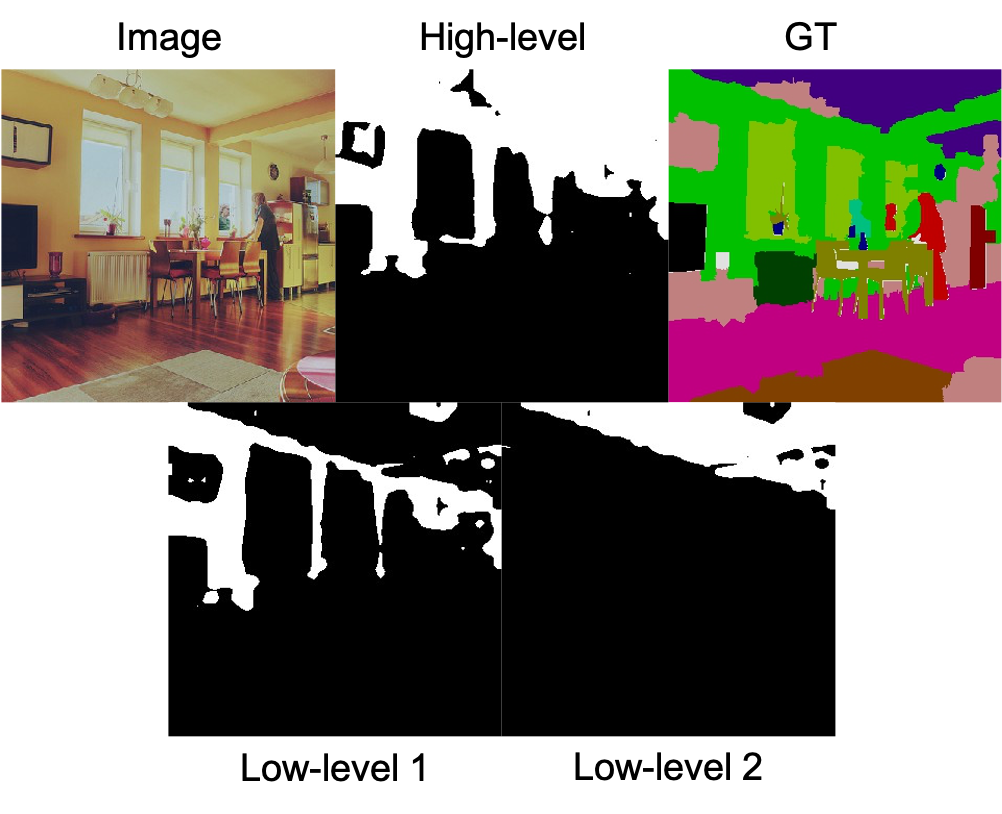} \\
     & 
\end{tabular}
\caption{{\bf Our hierarchy represents both similar classes and object parts.} In the top row, middle column, we visualize the higher-level cluster whereas the bottom row visualizes the corresponding lower-level ones. Our hierarchy can reason about part-whole relationships and class-class relationships.}
\label{fig:hier_qualt}
\end{figure*}

\subsection{Hierarchy as Structured Sparsity}
\label{structure}

In this section, we explain how we impose a structure between the clustering heads $\bm{H}^{(i)}$ from different levels $i$ of the hierarchy.
Specifically, we do so by introducing two operations: 1) Optimal Transportation between the cluster centers $\bm{H}^{(i)}$ and $\bm{H}^{(i+1)}$ to compute a soft but sparse relationship matrix $\bm{A}^{(i, i+1)} \in [0,1]^{\rho^i K \times \rho^{i+1} K}$ among the clusters; 2) matching the activations of $\bm{M}^{(i)}$ and $\bm{M}^{(i+1)}$ at each location for each image via a matching loss.

\textbf{Optimal Transportation for Hierarchy.} We formulate the hierarchy as a soft but sparse transportation matrix between the cluster centers $\bm{A}^{(i, i+1)} \in [0,1]^{\rho^i K \times \rho^{i+1} K}$ that ties the higher-level clusters to the lower-level ones. To obtain such a hierarchy between clusters $\bm{H}^{(i)}$ and $\bm{H}^{(i+1)}$, we solve an optimal transport problem~\cite{sinkhorn1967sinkhorn, cuturi2013sinkhornot}. The cost-per-mass matrix in this problem is modeled using the cosine distance between the cluster centers. Specifically, we compute the non-negative cosine similarity $\mathbf{Z}^{(i,i+1)}\in [0,1]^{\rho^i K \times \rho^{i+1} K}$ between the cluster centers $\bm{H}^{(i)}$ and $\bm{H}^{(i+1)}$ as
\begin{align}
    \mathbf{Z}^{(i,i+1)} = sim^+(\bm{H}^{(i)}, \bm{H}^{(i+1)})\;.
\end{align}
Our cost-per-mass matrix for the optimal transport problem is then computed as
\begin{align}
    \mathbf{C}^{(i,i+1)} = \mathbf{1} - \mathbf{Z}^{(i,i+1)}.
\end{align}
This cost-per-mass matrix imposes a higher transportation cost between the clusters whose centers are far from each other. Following this cost-per-mass matrix, we find the optimal transportation plan $\mathbf{A}^{(i,i+1)}$ by iteratively solving
\begin{align}
     \mathbf{A}^{(i,i+1)} &= \arg \min_{\mathbf{A}} \langle \mathbf{A}, \mathbf{C}^{(i,i+1)} \rangle_F + \frac{1}{\lambda} h(\mathbf{A})\\
    \text{s.t.} \quad & \mathbf{A} \mathbf{1}_{2^i K} = \mathbf{p}_y \; , \; \mathbf{A}^T \mathbf{1}_{2^{(i+1)} K} = \mathbf{p}_x
    \label{eq:ot}
\end{align}
where $h(\mathbf{A})$ denotes the entropy of $\mathbf{A}$, $\lambda$ is the entropy regularization parameter that controls the level of sparsity, $\mathbf{p}^{(i)} \in \mathbb{R}^{\rho^i K}$ and $\mathbf{p}^{(i+1)}$ are the marginal distributions that constrain the row and column sums of $\mathbf{A}^{(i,i+1)}$. They are chosen as uniform distributions to encourage balanced volumes for the clusters. We utilize Sinkhorn's algorithm~\cite{sinkhorn1967sinkhorn, cuturi2013sinkhornot} to solve this Optimal Transport problem. Sinkhorn's algorithm normalizes the sums of rows and columns of a matrix iteratively to yield a doubly stochastic matrix with desired row and column sums. It has been used in various computer vision applications~\cite{ozaydin2022dsi2i, liu2020scot, kolkin2019strotss, zhan2021unite, caron2020swav} to prevent many-to-many matches and empty clusters, and to encourage sparse matches. Nevertheless, we are the first to model the sparse hierarchy among clusters as an optimal transport problem in order to solve the USS problem.

\textbf{Wasserstein Loss for Hierarchy.} Once we find the sparse optimal transportation $\mathbf{A}^{(i,i+1)}$, we utilize it to increase the intersection of soft cluster assignments of matching pairs of clusters. Our aim is to allow intersection between clusters $k, l$ that have high values in $(\mathbf{A}^{(i,i+1)})_{k,l}$, where $k$ refers to the index of a higher-level cluster and $l$ indexes a lower-level cluster.

To that end, we match the activations of corresponding clusters from different levels, at each location. Specifically, we employ the $L_{\infty}$ Wasserstein Distance~\cite{villani2009otvillani, bogachev2012monge, cuturi2013sinkhornot} between the higher- and lower-level cluster activations. We first compute the joint cluster activation map $\bm{M}^{(i,i+1)} \in \{0,1\}^{\rho^i K \times \rho^{i+1} K \times HW}$ as
\begin{align}
    (\bm{M}^{(i,i+1)})_{k,l,p} &= (\bm{A}^{(i,i+1)})_{k,l} (\bm{M}^{(i+1)})_{l,p}\;.
\end{align}
Since $\bm{A}^{(i,i+1)}$ is a sparse matching matrix, only the joint activations for a few cluster pairs remain large after this multiplication. We then obtain transported higher-level activations by max-pooling over the lower-level cluster indices as
\begin{align}
    (\hat{\bm{M}}^{(i,i+1)})_{k,p} &= \max_l (\bm{M}^{(i,i+1)})_{k,l,p}\;.
\end{align}
Matching lower- and higher-level clusters via max-pooling has two motivations: 1) Matching each lower-level cluster with the single closest higher-level cluster at each point is intuitively desirable for a hierarchy; 2) max-pooling the joint distribution computed via an Optimal Transport matrix theoretically parallels the $L_{\infty}$ Wasserstein Distance~\cite{villani2009otvillani, bogachev2012monge, cuturi2013sinkhornot}. This lets us define the matching loss as
\begin{align}
    L^{{\text{match}}_{i,i+1}} = \sum_{x,k,p}{|(\hat{\bm{M}}_x^{(i,i+1)} - \bm{M}^{i}_x)_{k,p}|_1}\;.
    \label{eq:match}
\end{align}

Our matching loss transports the lower-level cluster activation at each location to the best matching higher-level cluster. This sparse matching with multiple levels of hierarchy provides a structure to the sparsity that is backpropagated to the features $F$ in Eq.~\ref{eq:sdh_cluster}. Hence, $L_\text{match}$ in Eq.~\ref{eq:match} provides a structure to the sparsity of $L_\text{cluster}$ in Eqs.~\ref{eq:stego_cluster},~\ref{eq:sdh_cluster}, which leads to an awareness of part-whole relationships and class-class relationships, and yields higher accuracy and mIoU for the USS task. Our ablation studies in the next section reveal that the structure of this sparsity is at the crux of the benefits of our method. Specifically, a structure imposed by multiple clustering heads ($N > 1$), an unequal number of clusters ($\rho > 1$), and a sparse hierarchy matrix (small $\lambda$ in Eq.~\ref{eq:ot}) are crucial for our method.

Our OMH can be incorporated in any of the previous USS frameworks~\cite{hamilton2022stego, seong2023hiddenpos, lan2023smooseg} during training time, without any additional test time complexity. During training, in addition to the original predictor and the original losses of the baselines, we introduce our hierarchical predictors $\bm{H}^{(i)}$, hierarchy matrices $\mathbf{A}^{(i,i+1)}$, and matching losses $L^{\text{match}}_{i,i+1}$ to minimize 
\begin{align}
    L = L^\text{base} + \lambda^{\text{sparsity}} \sum_{i}{L^\text{cluster}_i} + \lambda^{\text{structure}} \sum_{i}{L^{\text{match}}_{i,i+1}}
\end{align}
where $L^\text{base}$ refers to the original losses of the baselines. We then utilize the original predictor of each method, discarding the hierarchical predictors $\bm{H}^{(i)}$ and hierarchy matrices $\mathbf{A}^{(i,i+1)}$, to have exactly the same test setup as the previous works~\cite{hamilton2022stego, seong2023hiddenpos, lan2023smooseg}. This emphasizes the effectiveness and generalizability of our work. 
\section{Experiments}

\subsection{Datasets}

We follow the experimental setup of previous works \cite{lan2023smooseg, cho2021picie} and report our results on the COCOStuff27~\cite{caesar2018coco}, Cityscapes~\cite{Cordts2016cityscapes}, and Potsdam-3~\cite{ji2019iic} datasets. COCOStuff~\cite{caesar2018coco} consists of scene-centric images and has 91 stuff and 80 things categories. COCOStuff27~\cite{ji2019iic} is a version of COCOStuff where the 171 classes are merged into 27 mid-level classes~\cite{ji2019iic} as 15 stuff and 12 things categories. Cityscapes~\cite{Cordts2016cityscapes} is a street scene dataset consisting of 27 classes. Potsdam-3~\cite{ji2019iic} consists of aerial images with 3 classes. 

\subsection{Evaluation}

Following~\cite{ji2019iic, hamilton2022stego, seong2023hiddenpos, lan2023smooseg}, the predicted clusters are matched with the ground-truth classes through Hungarian matching. We utilize the original, non-hierarchical clustering predictor of each method, which makes our evaluation setting exactly the same as that of the baselines~\cite{hamilton2022stego, seong2023hiddenpos, lan2023smooseg}, wherein we maintain the same number of parameters, the number of clusters, and the evaluation architecture.

\subsection{Implementation Details}
We borrow the losses and hyperparameters from the baselines and use them unchanged. The $N$ clustering predictors of our OMH are trained with the Adam optimizer with the same hyperparameters as in the clustering probe of the baselines. The depth of the hierarchy is $N=3$, and the top level of the hierarchy has the same number of clusters $K$ as the number of classes. We set $\lambda^\text{sparsity}=0.01, \lambda^{\text{structure}}=0.3$.

For the experiments with STEGO~\cite{hamilton2022stego} and HP~\cite{seong2023hiddenpos}, we use the predictor of STEGO as $H^{(i)}$ for hierarchical clustering. For the experiments with SmooSeg~\cite{lan2023smooseg}, we use the predictor of SmooSeg as $H^{(i)}$. All the other losses and components of the baselines remain unchanged, including the stop gradient in their original predictors.

\subsection{Quantitative Results}
Our method outperforms recent baselines in terms of mIoU and Accuracy in Cityscapes, COCOStuff and Potsdam datasets. On Cityscapes dataset, in Table~\ref{Tab.cityscapes}, our method outperforms STEGO and SmooSeg in terms of Accuracy, in the reported metrics. In our experiments, we used 32-bit precision training for numerical stability with Sinkhorn's Algorithm~\cite{cuturi2013sinkhornot} but cannot reproduce SmooSeg in 32-bit precision. We outperform the best reproduction with the official codebase of SmooSeg in 32-bit training, in terms of both mIoU and Accuracy. In Tables~\ref{Tab.potsdam}~\ref{Tab.coco}, our method outperforms the baselines in terms of mIoU and Accuracy.

\begingroup
\setlength{\tabcolsep}{4.2pt} 
\renewcommand{\arraystretch}{1.0} 
\begin{table}[h]
    \centering
    \small
    \begin{tabular}{l c c c}
        \hline
        \multirow{2}{*}{Method} & \multirow{2}{*}{Backbone} &
        \multicolumn{2}{c}{Unsupervised} \\
        \multicolumn{2}{l}{} & Acc. & mIoU \\
        \hline
        \hline
        DINO~\cite{caron2021dino} & ViT-B/8 & 43.6 & 11.8 \\
        + STEGO~\cite{hamilton2022stego} & ViT-B/8 & 73.2 & 21.0 \\
        + OMH~(Ours) & ViT-B/8 & \textbf{74.8} & \textbf{21.9} \\ 
        \hline
        DINO \cite{caron2021dino} & ViT-S/8  & 40.5 & 13.7  \\
		+ TransFGU \cite{yin2022transfgu} & ViT-S/8  & 77.9 & 16.8 \\
		+ STEGO \cite{hamilton2022stego} & ViT-S/8 & 69.8 & 17.6  \\
		+ SmooSeg~\cite{lan2023smooseg} & ViT-S/8 & 82.8 & 18.4  \\
		+ SmooSeg* & ViT-S/8 & 80.9 & 17.7  \\
		+ OMH~(Ours) & ViT-S/8 & \textbf{83.5} & \textbf{18.2}  \\
        \hline
    \end{tabular}
    \caption{Experimental results on Cityscapes dataset. * denotes the best numbers reproduced by us.}
    \label{Tab.cityscapes}
\end{table}
\endgroup

\begingroup
\setlength{\tabcolsep}{6pt} 
\renewcommand{\arraystretch}{1.0} 
\begin{table}[!t]
    \centering
    \small
    \begin{tabular}{l c c}
        \hline
        Method & Backbone & Unsup. Acc. \\
        \hline
        DINO~\cite{caron2021dino} & ViT-B/8 & 53.0 \\
        + STEGO~\cite{hamilton2022stego}& ViT-B/8 & 77.0 \\
        + HP~\cite{seong2023hiddenpos} & ViT-B/8 & 82.4\\
        + OMH~(Ours) & ViT-B/8 & \textbf{82.7}\\
        \hline
    \end{tabular}
    \caption{Experimental results on Potsdam-3 dataset.}
    \label{Tab.potsdam}
\end{table}
\endgroup

\begingroup
\setlength{\tabcolsep}{4.2pt} 
\renewcommand{\arraystretch}{1.0} 
\begin{table}[ht]
    \centering
    \small
    \begin{tabular}{l c c c}
        \hline
        \multirow{2}{*}{Method}& \multirow{2}{*}{Backbone} & \multicolumn{2}{c}{Unsupervised} \\
        \multicolumn{2}{l}{}  & Acc. & mIoU \\
        \hline
        \hline
        DINO~\cite{caron2021dino} & ViT-S/8 & 28.7 & 11.3 \\ 
        + TransFGU~\cite{yin2022transfgu} & ViT-S/8 & 52.7 & 17.5 \\
        + STEGO~\cite{hamilton2022stego} & ViT-S/8 & 48.3 & 24.5 \\
        + HP~\cite{seong2023hiddenpos} & ViT-S/8 & 57.2 & 24.6 \\ 
        + OMH~(Ours) & ViT-S/8 & \textbf{57.6} & \textbf{25.3} \\ 
        \hline
        DINO~\cite{caron2021dino} & ViT-B/8 & 30.5 & 9.6 \\ 
        + STEGO~\cite{hamilton2022stego} & ViT-B/8 & 56.9 & 28.2 \\ 
        + OMH~(Ours) & ViT-B/8 & \textbf{57.7} & \textbf{30.1} \\ 
        \hline
    \end{tabular}
    \caption{Experimental results on COCO-stuff dataset with various backbones and pre-trained models.}
    \label{Tab.coco}
\end{table}
\endgroup

\begin{figure*}[ht]
    \includegraphics[width=0.97\linewidth]{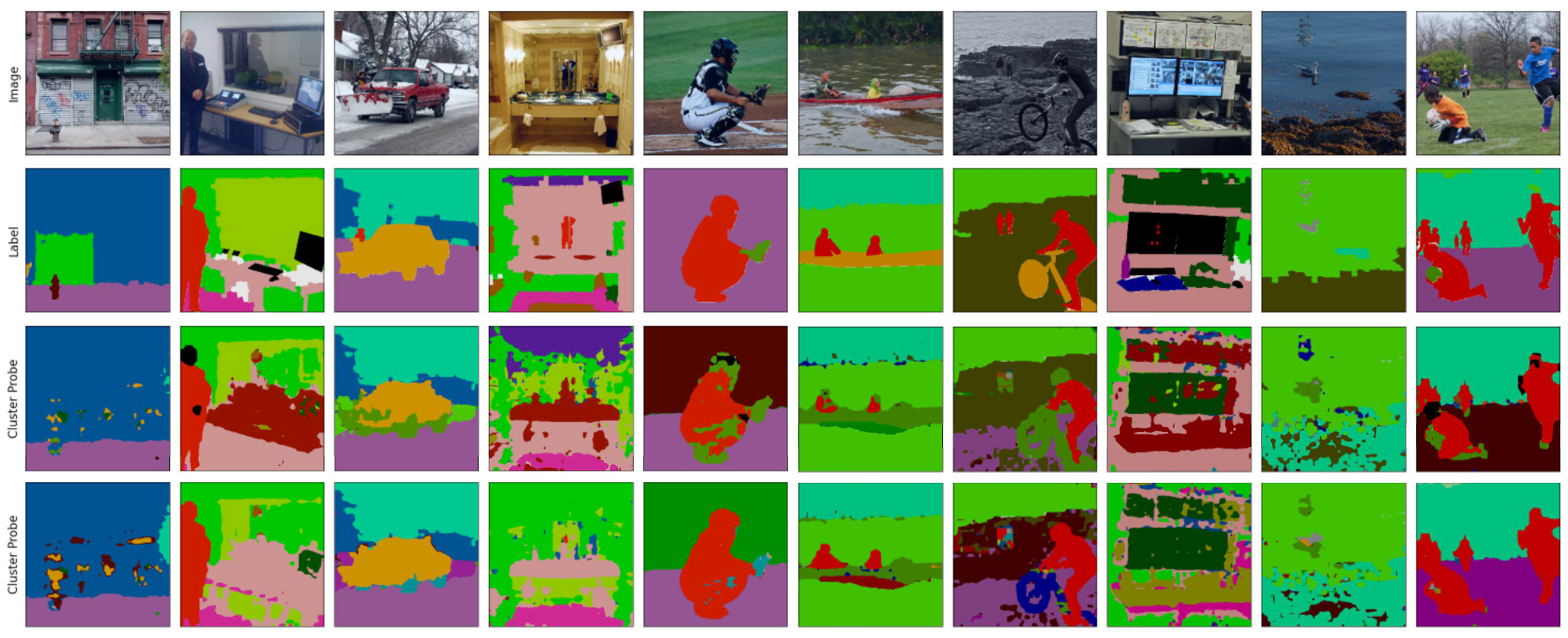}
\caption{{\bf Qualitative results for OMH.} Qualitative results demonstrating the contribution of our method. The third row is without OMH and the fourth one is with OMH. Results from Table~\ref{Tab.coco} last line, without CRF.}
\label{fig:qualt}
\end{figure*}

\subsection{Ablation Study}
We begin our ablation study by analyzing the interplay between $\lambda^\text{sparsity}, L^\text{cluster}$ and $\lambda^{\text{structure}}, L^\text{match}$. In Table \ref{Tab.abl_loss}, we observe that structure without sparsity or sparsity without structure are not as beneficial as their combined effects, structured sparsity. We then analyze the structure of our OMH in Tables~\ref{Tab.abl_levels}, \ref{Tab.abl_expansion}, and~\ref{Tab.abl_temp}. 

\textbf{Levels of the hierarchy.} Table~\ref{Tab.abl_levels} shows that having multiple levels in our hierarchy is beneficial. In most SSL and USS frameworks, a two-level, siamese architecture seems effective~\cite{chen2021simsiam,he2020mocov1,chen2020simclr}. Contrary to these works, having three levels of the hierarchy seems more effective for our framework, compared to having two.

\textbf{Expansion factor.}  In Table~\ref{Tab.abl_expansion}, we observe that having unequal numbers of clusters in different levels of the hierarchy is beneficial. Note that an expansion factor of 1 with 2 levels of hierarchy imitates the previous SSL frameworks with a siamese architecture~\cite{chen2021simsiam,he2020mocov1,chen2020simclr}. However, our method benefits from having unequal numbers of clusters, which enables the hierarchy to focus on different levels of details (higher-level vs lower-level).

\textbf{OT Temperature.} Having a sparse, few-to-few relationship across the clusters from different levels is crucial for the performance of our method. We observe in Table~\ref{Tab.abl_temp} that a larger entropy regularization coefficient $\lambda$ in Eq.~\ref{eq:ot} leads to a larger entropy, less sparsity, and less accuracy for the USS task. We use the smallest possible $\lambda=0.02$ in our experiments with Sinkhorn's algorithm~\cite{sinkhorn1967sinkhorn,cuturi2013sinkhornot}, as $\lambda<0.02$ leads to NaN in our case, with 32-bit precision. 

The entropy regularization term $\lambda$ in our OMH can be linked to the temperature in the seminal work on Knowledge Distillation~\cite{hinton2015distilling}. In Knowledge Distillation~\cite{hinton2015distilling}, the noisy predictions of a teacher model, with a lower softmax temperature, are useful for training a student model. In our case, the parallel clusterings in our hierarchy are tied to each other with a low temperature via OT, and may act as teachers for each other.


\begingroup
\setlength{\tabcolsep}{4.2pt} 
\renewcommand{\arraystretch}{1.0} 
\begin{table}[ht]
    \centering
    \small
    \resizebox{\columnwidth}{!}{
    \begin{tabular}{l c c c}
        \hline
        \multirow{2}{*}{Method}& \multirow{2}{*}{Backbone} & \multicolumn{2}{c}{Unsupervised} \\
        \multicolumn{2}{l}{}  & Acc. & mIoU \\
        \hline
        STEGO~\cite{hamilton2022stego} & ViT-B/8 & 56.9 & 28.2 \\ 
        + $L^\text{cluster}$ & ViT-B/8 & 56.5 & 27.9 \\
        + $L^\text{match}$ & ViT-B/8 & \textbf{57.7} & \textbf{30.1} \\
        \hline
    \end{tabular}}
    \caption{Our matching loss is at the core of our contribution.}
    \label{Tab.abl_loss}
    \vspace{-0.1cm}
\end{table}
\endgroup

\begingroup
\setlength{\tabcolsep}{4.2pt} 
\renewcommand{\arraystretch}{1.0} 
\begin{table}[ht]
    \centering
    \small
    \resizebox{\columnwidth}{!}{
    \begin{tabular}{l c c c }
        \hline
        \multirow{2}{*}{Method}& \multirow{2}{*}{Backbone} & \multicolumn{2}{c}{Unsupervised} \\
        \multicolumn{2}{l}{}  & Acc. & mIoU \\
        \hline
        STEGO~\cite{hamilton2022stego} & ViT-B/8 & 56.9 & 28.2 \\ 
        OMH~/w 1 hierarchy (Ours w/o $L^\text{match}$) & ViT-B/8 & 56.5 & 27.9 \\
        OMH~/w 2 hierarchy & ViT-B/8 & 57.3 & 29.0 \\
        OMH~/w 3 hierarchy & ViT-B/8 & \textbf{57.7} & \textbf{30.1} \\
        OMH~/w 4 hierarchy & ViT-B/8 & 56.9 & 28.5 \\
        \hline
    \end{tabular}}
    \caption{\textbf{Number of levels in the hierarchy.} Structured sparsity via a 3-level hierarchy leads to the best performance. Sparsity without a structure (line 2) is not beneficial for the USS task.}
    \label{Tab.abl_levels}
    \vspace{-0.3cm}
\end{table}
\endgroup

\begingroup
\setlength{\tabcolsep}{4.2pt} 
\renewcommand{\arraystretch}{1.0} 
\begin{table}[ht]
    \centering
    \small
    \resizebox{\columnwidth}{!}{
    \begin{tabular}{l c c c }
        \hline
        \multirow{2}{*}{Method}& \multirow{2}{*}{Backbone} & \multicolumn{2}{c}{Unsupervised} \\
        \multicolumn{2}{l}{}  & Acc. & mIoU \\
        \hline
        STEGO~\cite{hamilton2022stego} & ViT-B/8 & 56.9 & 28.2 \\ 
        OMH~/w expansion 1 & ViT-B/8 & 57.2 & 28.7 \\
        OMH~/w expansion 1.5 & ViT-B/8 & 57.6 & 30.1 \\
        OMH~/w expansion 2 & ViT-B/8 & \textbf{57.7} & \textbf{30.1} \\
        OMH~/w expansion 3 & ViT-B/8 & 56.1 & 27.4 \\
        \hline
    \end{tabular}}
    \caption{\textbf{Expansion factor.} Ratio of the number of clusters for two consecutive levels. Expanding the number of clusters by a factor of 1.5-2 leads to the best performance.}
    \label{Tab.abl_expansion}
    \vspace{-0.4cm}
\end{table}
\endgroup

\begingroup
\setlength{\tabcolsep}{4.2pt} 
\renewcommand{\arraystretch}{1.0} 
\begin{table}[ht]
    \centering
    \small
    \resizebox{\columnwidth}{!}{
    \begin{tabular}{l c c c }
        \hline
        \multirow{2}{*}{Method}& \multirow{2}{*}{Backbone} & \multicolumn{2}{c}{Unsupervised} \\
        \multicolumn{2}{l}{}  & Acc. & mIoU \\
        \hline
        STEGO~\cite{hamilton2022stego} & ViT-B/8 & 56.9 & 28.2 \\ 
        OMH~/w temp 0.02 & ViT-B/8 & \textbf{57.7} & \textbf{30.1} \\
        OMH~/w temp 0.05 & ViT-B/8 & 57.1 & 29.8 \\
        OMH~/w temp 0.10 & ViT-B/8 & 56.5 & 28.5 \\
        \hline
    \end{tabular}}
    \caption{\textbf{Sparsity.} A low $\lambda$ that leads to sparser solutions to the Optimal Transport problem is at the crux of our method. We use the smallest $\lambda$ (0.02) that is computationally stable for OT.}
    \label{Tab.abl_temp}
    \vspace{-0.4cm}
\end{table}
\endgroup
\section{Conclusion}
We have argued that semantic sparsity and lack of clear class definitions need to be explicitly addressed in the USS task. Hence, we have proposed our Optimally Matched Hierarchy to impose a hierarchical structure and sparsity on the learned features. Our experiments on three datasets with two backbones and three USS methods have shown that the structured sparsity induced via our OMH is beneficial for unsupervised segmentation.

\section{Limitations and Future Work}
Even though the hierarchical clusters found by OMH are useful for backpropagating structured sparsity to the features in USS, the outputs of our hierarchy are not exploited explicitly. We hope that our work will spearhead future research in the development of hierarchical representations for unsupervised learning.

\section{Impact Statement}
This paper presents work whose goal is to advance the field of Machine Learning. There are many potential societal consequences of our work, none of which we feel must be specifically highlighted here.

\textbf{Acknowledgement.} This work was supported
by the Swiss National Science Foundation via the Sinergia
grant CRSII5-180359. 

\newpage
\bibliography{main}
\bibliographystyle{icml2024}

\newpage
\appendix
\onecolumn

\end{document}